\documentclass[accepted]{uai2021} 

\usepackage[american]{babel}

\usepackage{natbib} 
\usepackage{mathtools} 
\usepackage{booktabs} 
\usepackage{tikz} 

\title{Towards Robust Object Detection: Bayesian RetinaNet for Homoscedastic Aleatoric Uncertainty Modeling}

\author[1]{\href{mailto:<nehanzhina@itmo.ru>?Subject=Your UAI 2021 paper}{Natalia~Khanzhina}{}}
\author[1]{Alexey~Lapenok}
\author[1]{Andrey~Filchenkov}
\affil[1]{%
Machine Learning Lab\\
ITMO University\\
49 Kronverksky Pr., St. Petersburg 197101, Russia
}
  
\begin{document}
\maketitle

\begin{abstract}
According to recent studies, commonly used computer vision datasets contain about 4\% of label errors. For example, the COCO dataset is known for its high level of noise in data labels, which limits its use for training robust neural deep architectures in a real-world scenario.

To model such a noise, in this paper we have proposed the homoscedastic aleatoric uncertainty estimation, and present a series of novel loss functions to address the problem of image object detection at scale.

Specifically, the proposed functions are based on Bayesian inference and we have incorporated them into the common community-adopted object detection deep learning architecture~RetinaNet.

We have also shown that modeling of homoscedastic aleatoric uncertainty using our novel functions allows to increase the model interpretability and to improve the object detection performance being evaluated on the COCO dataset.


  
\end{abstract}

\section{Introduction}\label{sec:intro}

Usually, training a predictive algorithm involves training a machine learning model on a labeled dataset from a scratch or using this dataset to fine-tune a model previously pre-trained on a large publicly available dataset such as ImageNet or MS COCO. 
However, a recent study~\citep{errors} concluded that commonly used open datasets for computer vision tasks contain about 4\% of errors in image labels. The MS COCO dataset for detection models benchmarking is also known for its noisy labels of both object classes and bounding boxes~\citep{khetan2017learning,vahdat2017toward}.
At the same time, popular cross-entropy loss is considered to be sensitive to noisy labeling~\citep{feng2020can}.
Moreover, the deeper the model, the more it adapts to these labeling errors. 
This negatively affects not only the integrity of the contests on the corresponding datasets, but also a real-world scale, since these datasets are often used for model pre-training to solve various problems.

One way to account for the label errors is to estimate the aleatoric uncertainty, which reflects the noise level in the training data and can be used at the inference time~\citep{hullermeier2021aleatoric}. 
The aleatoric uncertainty is divided into homoscedastic, i.e. constant for the data distribution in a particular task, and heteroscedastic, i.e. different for each data object~\citep{uncertainties}. 
Despite the estimation of heteroscedastic uncertainty is more useful for computer vision problems in general~\citep{uncertainties}, its modeling requires changes in the neural network (NN) architecture.
Moreover, its application in practice requires developing tools to postprocess prediction for a particular object with this uncertainty.

At the same time, the modeling of homoscedastic aleatoric uncertainty can be performed based on the modification of the loss functions rather than the architecture, which is less time-consuming. 
In addition, homoscedastic aleatoric modeling even improves the accuracy of solving the computer vision problems~\citep{yarin-gal}.  
Researchers~\citet{yarin-gal} consider the application of modeling this type of uncertainty for multi-task NN architecture, solving semantic, instance segmentation, and depth regression problems. 
Quantification of aleatoric uncertainty can greatly increase model performance in the detection problem~\citep{feng2019leveraging,meyer2019lasernet}.

Recently, Bayesian deep learning has been widely used in object detection
\citep{bendale2016towards,harakeh2020bayesod,kraus2019uncertainty,miller2018dropout,miller2019evaluating,miller2021uncertainty,postels2019sampling}.
However, all these works focus on epistemic uncertainty.

Fewer number of papers are devoted to the aleatoric uncertainty estimation~\citep{kraus2019uncertainty,le2018uncertainty}
including those on 3d object detection~\citep{feng2018towards,feng2019leveraging,meyer2019lasernet} and one-stage detector~\citep{kraus2019uncertainty,le2018uncertainty}.
However, existing works do not study the application of homoscedastic aleatoric uncertainty modeling for the detection problem, although this can help isolate noise from data and improve model robustness. Moreover, as the detection is the multi-task problem (i.e. includes localization and classification tasks), the modeling can be performed without changes in the neural network architecture, using tools, developed by~\citet{yarin-gal}.

Being inspired by this, we aimed to answer the following research questions:

{\bf RQ1}: Can homoscedastic aleatoric uncertainty modeling improve the detection accuracy based on deep neural networks?  

{\bf RQ2}: Can Bayesian approximation be effectively applied to modeling homoscedastic aleatoric uncertainty for existing detection models?

In order to answer them, we propose novel loss functions, whose optimization is equivalent to modeling homoscedastic aleatoric uncertainty for the joint localization and classification tasks. 
The paper contributions are the following:
\begin{enumerate}
\item A new loss function for the classification task for modeling the aleatoric uncertainty called \textbf{Bayesian Focal Loss}.
\item A new loss function for the localization task for modeling the aleatoric uncertainty called \textbf{Bayesian Smooth} $\mathbf{L_1}$ \textbf{Loss}.
\end{enumerate}

The proposed loss functions for modeling the homoscedastic aleatoric uncertainty can be applied to any NN detectors, which use cross-entropy or Focal loss and $L_1$ or Smooth $L_1$ loss, without changing their architecture and training pipeline. 
The uncertainty modeling can make existing detectors robust to noise in data labels and can improve detection accuracy as well.

\section{Related Work}\label{sec:rl}
\subsection{Bayesian Deep Learning for computer vision}\label{sec:bdl}

Recently,~\citet{yarin-gal} suggested a tool for modeling homoscedastic aleatoric uncertainty to weigh multi-task losses. 
They considered three computer vision tasks: semantic segmentation, instance segmentation, and depth regression. 
The modeling required building a probabilistic model for both classification and regression tasks.

For the regression task, they defined a probabilistic model with a Gaussian likelihood, where the mean is given by the model output $f^W(x)$ with weights $W$ on input $x$:
\begin{equation}
    p\left(y|f^W(x)\right) = N(f^W(x),\sigma^2)
\end{equation}
and the variance is given as an observation noise scalar $\sigma$, which captures homoscedastic aleatoric uncertainty. 

Interpreting this Gaussian log likelihood maximization as objective, they obtained the modification of $L_2$ loss:
\begin{equation}
    BL_2=-\dfrac{1}{2\sigma^2}\left\|y-f^W(x)\right\|^2 - \log\sigma,
\end{equation}
with $BL_2$ being the Bayesian $L_2$ loss. It is then maximized with respect to weights $W$ and noise scalar $\sigma$.

For the classification task, the likelihood appeared to be less trivial. Assume the model output $f^W(x)$ is scaled by ${1}/{\sigma^2}$ and then squashed through the Softmax activation function. Then, the likelihood is the following:
\begin{equation}
    p\left(y|f^W(x),\sigma\right)=Softmax\left(\dfrac{1}{\sigma^2}f^W(x)\right),
\end{equation}
which can be interpreted as the Boltzmann distribution with temperature $\sigma$.

The log likelihood is defined as:
\begin{equation}
\begin{aligned}
    \log p\left(y=c|f^W(x),\sigma\right)=\dfrac{1}{\sigma^2}f_c^W(x) - \\
    - \log  \sum_{c'}{\exp\left(\dfrac{1}{\sigma^2}f_{c'}^W(x)\right)},
\end{aligned}\label{bsoftmax}
\end{equation}
with $f_c^W(x)$ the element of $f^W(x)$ vector for a particular class $c$.

Using maximum likelihood inference for the multi-task neural network with output $y_1$ for the regression task and $y_2$ for the classification task the following minimization objective can be obtained:
\begin{equation}
\begin{aligned}
    L(W, \sigma_1,\sigma_2)&=\dfrac{1}{2\sigma_1^2}L_1(W)+\log\sigma_1+\\
    &+\dfrac{1}{\sigma_2^2}L_2(W)+\log\sigma_2,
\end{aligned}
\end{equation}
where $L_1(W)=\|y_1-f^W(x)\|^2$ is the Euclidean loss for $y_1$ and $L_2(W)=-\log\left(Softmax(y_2,f^W(x)\right)$ is the cross-entropy loss for $y_2$. This loss is optimised with respect to $W$ as well as $\sigma_1$ and $\sigma_2$.

The main difficulty with the $L_2(W)$ loss is to release $x$ in $f^W(x)$ 
from scaling factor ${1}/{\sigma^2}$. To achieve this, \citet{yarin-gal} performed the following: subtracted and added $\dfrac{1}{\sigma^2} \log \left(\sum_{c'}\exp{(f_{c'}^W(x))}\right)$ to Eq.~\ref{bsoftmax}, then used a simplifying assumption 
$$\left[\sum_{c'}\exp\left(f^W_{c'}(x)\right)\right]^\frac{1}{\sigma^2} \approx \dfrac{1}{\sigma}\sum_{c'}\exp\left(\dfrac{1}{\sigma^2}f^W_{c'}(x)\right),$$ 
which becomes an equality when $\sigma\to 1$.
\subsection{RetinaNet detector}
RetinaNet~\citep{focal-loss} is a one-stage anchor-based neural network for object detection. 
This architecture is most famous by the proposed classification loss function, referred as Focal Loss. 
RetinaNet consists of four subnetworks:
\begin{itemize}
    \item Backbone is a basic convolutional network that extracts features from the input image. 
    Traditionally, the state-of-the-art networks are used as backbones, such as ResNet~\citep{resnet}, VGG~\citep{vgg}, EfficientNet~\citep{efficientNet}.
    \item Feature Pyramid Network (FPN) is a ``neck'' convolutional neural network proposed by~\citet{fpn}. 
    It combines feature maps from different layers of the backbone network in a top-down pathway using lateral connection. 
    This allows to solve a task (classification or regression) at different image resolutions and semantic scales.
    \item Localization subnetwork is a ``head'' subnetwork that extracts information from the FPN about the coordinates of objects in the image, solving the regression task. 
    It trains based on the Smooth $L_1$ loss proposed by~\citet{smooth-l1}.
    \item Classification subnetwork is a ``head'' subnetwork that extracts information about object classes from the FPN, solving the classification task. It trains based on the Focal loss.
\end{itemize}

For the bounding boxes regression, RetinaNet uses Smooth $L_1$ loss.
This is a combination of $L_1$ and $L_2$ loss functions, which was initially inspired by~\citet{huber1992robust}. 
Its formula is
\begin{equation}
Smooth_{L1}(x)=\begin{cases}
    \dfrac{\beta^2}{2} \cdot \epsilon^2, & \text{if} \ \epsilon<\dfrac{1}{\beta^2},\\
    \epsilon - \dfrac{1}{2 \beta^2}, & \text{otherwise}
\end{cases}
\end{equation} 
with ${1}/{\beta^2}$ the threshold for switching from the $L_1$ to the $L_2$ loss function, and $\epsilon=\|y - f^W(x)\|$ with $x$ the network input, its output $f^W(x)$, and $y$ the ground truth coordinate of the object bounding box. The main difference from the $L_2$ loss function is that addition of $L_1$ case helps avoid over-penalizing outliers. 

For the classification, \citet{focal-loss} introduced Focal loss. Focal loss is proven to penalize the network better than the cross-entropy loss~\citep{focal-loss} on hard negative examples. Its formula is
\begin{equation}
FL(p_t) = - \log p_t \cdot \left(1 - p_t\right)^\gamma,
\end{equation}
where 
$p_t=
\begin{cases} 
    p & \quad y=1, \\
    1 - p & \quad \text{otherwise}
\end{cases}$

with $p = Sigmoid(f^W(x))$, $y$ the ground truth class label of an object. The main difference of Focal loss from the cross-entropy loss is the modulationg factor $\gamma\in \left(0, +\infty\right)$ introduced to handle the problem of class imbalance, which is typical for object detection, since an object of interest usually occupies relatively little space in the image. 
Thus, Focal loss results in higher gradient values for higher error values and vice versa. This forces the network to focus on hard negative examples better, which are the objects of interest. 
The generalized RetinaNet loss function can then be written as 
$L = \alpha L_{class} + L_{reg}$, with $L_{class}$ the classification Focal loss function, $L_{reg}$ the regression (localization) Smooth $L_1$ loss function, $\alpha$ the balancing coefficient that adjusts the impact of the $L_{class}$ term.

Although RetinaNet loss functions are quite effective, they do not allow to capture homoscedastic aleatoric uncertainty making RetinaNet sensitive to the noisy data. To overcome this issue, we propose the novel Focal and Smooth $L_1$ loss functions, which are able to model homoscedastic aleatoric uncertainty. We call our neural network, that utilizes them, \textit{Bayesian RetinaNet}.

\section{Bayesian RetinaNet}\label{sec:method}
In this section, we introduce the novel loss functions with homoscedastic uncertainty based on maximum likelihood estimation.

Let $f^W(x)$ denote the output of a neural network with weights $W$ on input $x$ and $\epsilon$ be the error that is the norm of difference between the ground truth value and our prediction:
$$\epsilon=\|y - f^W(x)\|.$$
\subsection{Bayesian Smooth $L_1$ Loss for Homoscedastic Aleatoric Uncertainty}\label{sec:bl1}

First, we introduce the novel likelihood for the localization task, which is to predict object coordinates. As localization is the regression task, we adopt the likelihood from Section~\ref{sec:bdl} for the Smooth $L_1$ loss and define our likelihood as the combination of Gaussian and Laplace likelihoods:
\begin{equation}
     p\left(y|f^W(x), \sigma, \alpha \right) = \begin{cases}
            p_{G}\left(y|f^W(x),\sigma\right), & \text{if } \epsilon < \dfrac{1}{\beta^2}\\
            p_{L}\left(y|f^W(x),\alpha\right), & \text{otherwise}
        \end{cases},
\end{equation}
where $p_{G}$ is Gaussian likelihood, $p_{L}$ is Laplace likelihood with observation noise scalars $\sigma$ and $\alpha$, respectively. 

As in maximum likelihood inference, here we maximise the log likelihood of the model. Thus, following the likelihood for regression in the case of $L_2$ loss~\citep{yarin-gal}, for Smooth $L_1$ it can be written as
\begin{equation}
    \log p \left(y|f^W(x), \sigma, \alpha \right) \propto \begin{cases}
            -\dfrac{\epsilon^2}{2\sigma^2} - \log\sigma, & \text{if} \ \epsilon < \dfrac{1}{\beta^2}\\
            -\alpha\epsilon + \log\alpha \label{eq2:smooth_l1_base}& \text{otherwise}
    \end{cases},
\end{equation}
where $L_2$ corresponds to Gaussian likelihood $p_{G}$, $L_1$ to Laplace likelihood $p_{L}$.

This leads to the following minimization objective $L(W, \sigma, \alpha)$:
\begin{align}
    \begin{split}
        & L(W, \sigma, \alpha)= -\log p\left(y|f^W(x), \sigma\right) \propto \\
        &\propto \begin{cases}
            \dfrac{\epsilon^2}{2\sigma^2} + \log\sigma, & \text{if } \ \epsilon < \dfrac{1}{\beta^2},\\
            \alpha\epsilon - \log\alpha & \text{otherwise}
        \end{cases} \\
        &=\begin{cases}
            \dfrac{1}{2\sigma^2}L_2(W) + \log\sigma,& \text{if} \ \epsilon < \dfrac{1}{\beta^2},\\
            \alpha L_1(W)-\log\alpha & \text{otherwise}
        \end{cases} 
    \end{split}\label{eq2:smooth_l1_base}
\end{align}
where we write $L_1(W)=\epsilon$ for the $L_1$ loss of $y$, write $L_2(W)=\epsilon^2$ for Euclidean loss of $y$.

The likelihood in Eq.~\ref{eq2:smooth_l1_base} has two variances corresponding to $L_1$ and $L_2$. However, this is inconvenient in practice because the ground truth bounding box coordinates are unknown in the real world model inference. Thus, to find the dependency between $\alpha$ and $\sigma$ and also save the property of the likelihood, we solve the equation of density function of our likelihood:
\begin{equation}\label{eq2:smooth_l1_density}
    \int\limits_{-\infty}^{\infty} p(y|f^W(x), \sigma)dt = 1.
\end{equation}

From this equation, we obtain the following dependency between variances:
\begin{equation}\label{eq2:smooth_l1_alpha}
        \alpha = -\beta^2\log\tau,
\end{equation}
where $\tau=1 - erf{\left(\dfrac{1}{\beta^2\sqrt{2\sigma^2}}\right)}$ with $erf$ the Gauss error function~\citep{abramowitz1988handbook}.

Whether we place Eq.~\ref{eq2:smooth_l1_alpha} into Eq.~\ref{eq2:smooth_l1_base}, we obtain Bayesian Smooth $L_1$ Loss:
\begin{align}\label{eq2:smooth_l1_final}
    \begin{split}
    & BSmooth_{L1}(\epsilon) = \\
    & = \begin{cases}
        \dfrac{\epsilon^2}{2\sigma^2} + \log\sigma & \text{if} \ \epsilon < \dfrac{1}{\beta^2}, \\
         -\beta^2\epsilon\log \tau  - \log\left(-\beta^2\log \tau\right)&
         \text{otherwise}
    \end{cases}
    \end{split}
\end{align}

The first and second cases of Eq.~\ref{eq2:smooth_l1_final} are not equal, when $\epsilon$ equals to ${1}/{\beta^2}$. The second case requires a small correction. To solve this issue, we smooth this function and obtain the following loss function:
\begin{equation}
    \begin{aligned}
    & BSmooth_{L1}(\epsilon) = \\
    & = \begin{cases}
     \dfrac{\epsilon^2}{2 \sigma^2} + \log\sigma & \text{if} \ \epsilon < \dfrac{1}{\beta^2}\\
     -\beta^2\epsilon\log\tau + \log \tau +\dfrac{1}{2\sigma^2\beta^4}+\log\sigma
      & \text{otherwise}
    \end{cases}.
    \end{aligned}
\end{equation}

\begin{figure}[h]
    \centering
    \includegraphics[scale=0.3]{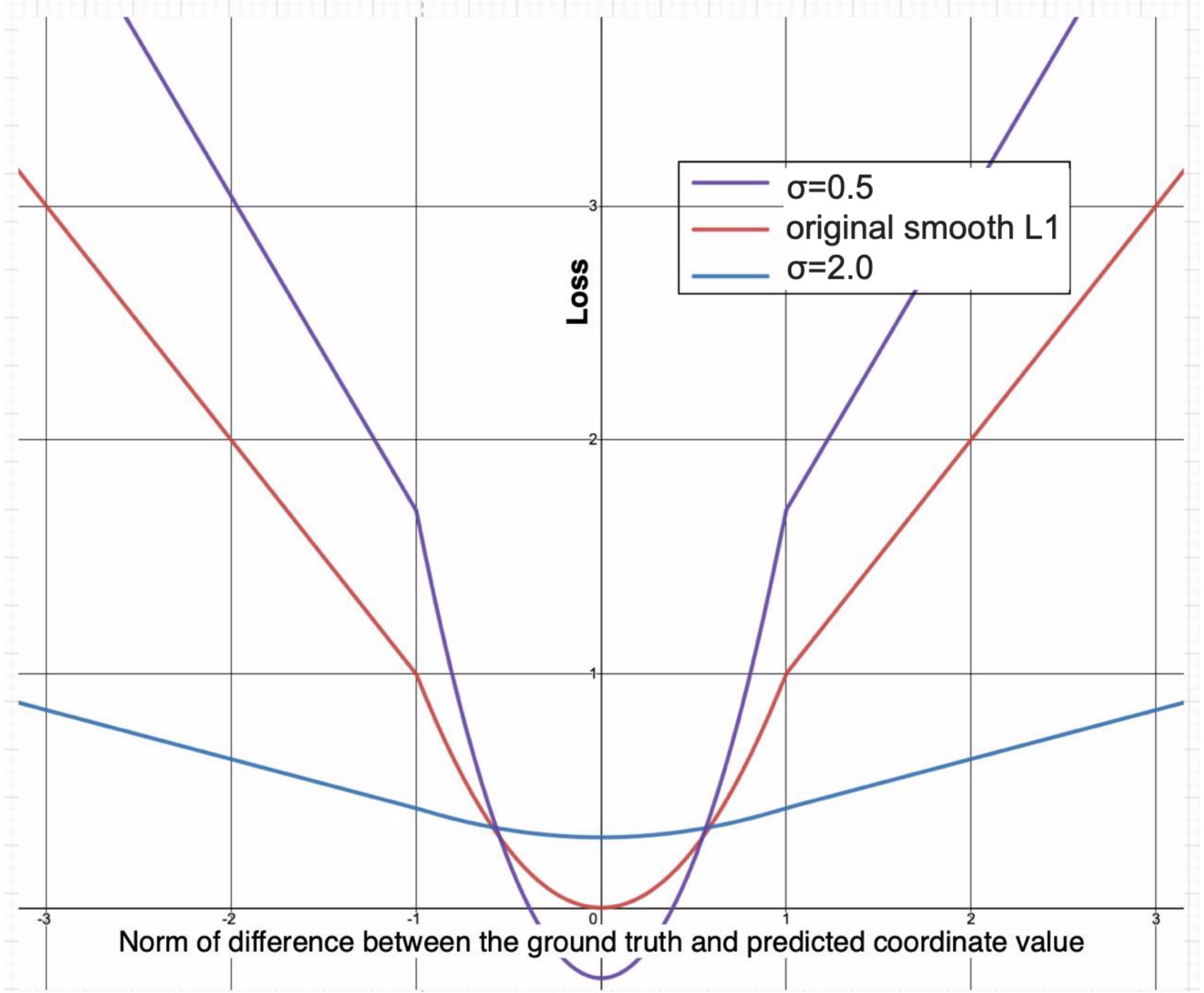}
    \caption{The proposed Bayesian Smooth $L_1$ loss function for different estimates of aleatoric uncertainty. At $\sigma=0.5$, $\sigma=2.0$, compared to the original Smooth $L_1$ loss (red line).}\label{fig:bl1}
\end{figure}

Following~\citet{yarin-gal} in experiments we train the network to predict the log variance, $s:=\log\sigma^2$, which is more numerically stable than regressing the variance $\sigma^2$ directly to avoid division by zero.
The proposed Bayesian Smooth $L_1$ loss function plot is presented in Fig.~\ref{fig:bl1} in comparison with the original Smooth $L_1$ loss. The proposed loss function penalizes the neural network better than the original Smooth $L_1$ loss: for less noisy data, it penalizes the neural network more for large prediction errors.
For noisier data, it penalizes the neural network more uniformly, less ``trusting'' the data labels.

\subsection{Bayesian Focal Loss for Homoscedastic Aleatoric Uncertainty}\label{sec:bfl}
Now we introduce the novel likelihood function for the classification task, which is the modified Focal loss. In RetinaNet, the classification activation function is logistic, which is more convenient for datasets with non-mutually exclusive classes.
Thus, for the classification task, the likelihood can be defined as:
\begin{equation}
         p\left(y|f^W(x),\sigma\right) = Sigmoid\left(\dfrac{1}{\sigma^2}f^W(x)\right)
 \end{equation}
with a positive noise scalar $\sigma$, which reflects homoscedastic uncertainty. This likelihood can also be interpreted as the Boltzmann distribution where the input is scaled by $\sigma^2$. 
We aim to maximise the likelihood. For the classification task it can be effectively done using Focal loss, which behaves the same way as log likelihood. Focal loss likelihood can be defined as:
\begin{equation}
\begin{aligned}
    BFL\left(p(y|f^W(x),\sigma\right)=-\left(
            \dfrac{1}{\sigma}\left(1-p_t\right)^{\sigma^{-2}}
        \right)^\gamma\times\\
        \times\left(\dfrac{1}{\sigma^2}\log{p_t} - \log\sigma\right)
\end{aligned}
\end{equation}
where BFL is Bayesian Focal loss and
$$p_t=\begin{cases}
    p & \quad y=1, \\
    1 - p & \quad \text{otherwise}
\end{cases}$$

To obtain BFL from the original Focal loss, the main issue is to release $f^W(x)$ in logistic function from the scaling factor ${1}/{\sigma^2}$. To solve this issue and obtain the new form of likelihood, the following transitions are used: subtraction and addition of $\dfrac{1}{\sigma^2}\log\left(1+\exp{\left(f^W(x)\right)}\right)$ term; simplifying assumptions
\begin{equation}
\begin{aligned}
\dfrac{1}{\sigma}\left[1+\exp{\left(\dfrac{1}{\sigma^2} f^W(x)\right)}\right]\approx \\
\approx\left[1+\exp{\left(f^W(x)\right)}\right]^{\dfrac{1}{\sigma^2}},
\end{aligned}
\end{equation} 
and
\begin{equation}
\begin{aligned}
\dfrac{1}{\sigma}\left[1+\exp{\left(-\dfrac{1}{\sigma^2} f^W(x)\right)}\right]\approx \\
\approx\left[1+\exp{\left(-f^W(x)\right)}\right]^{\dfrac{1}{\sigma^2}},
\end{aligned}
\end{equation}
which become an equality when $\sigma\to 1$.

Bayesian Focal loss is equal to the original Focal Loss, when $\sigma=1$, or $p(y|f^W(x)=Sigmoid(f^W(x))$.

\begin{figure}[h]
    \centering
    \includegraphics[scale=0.3]{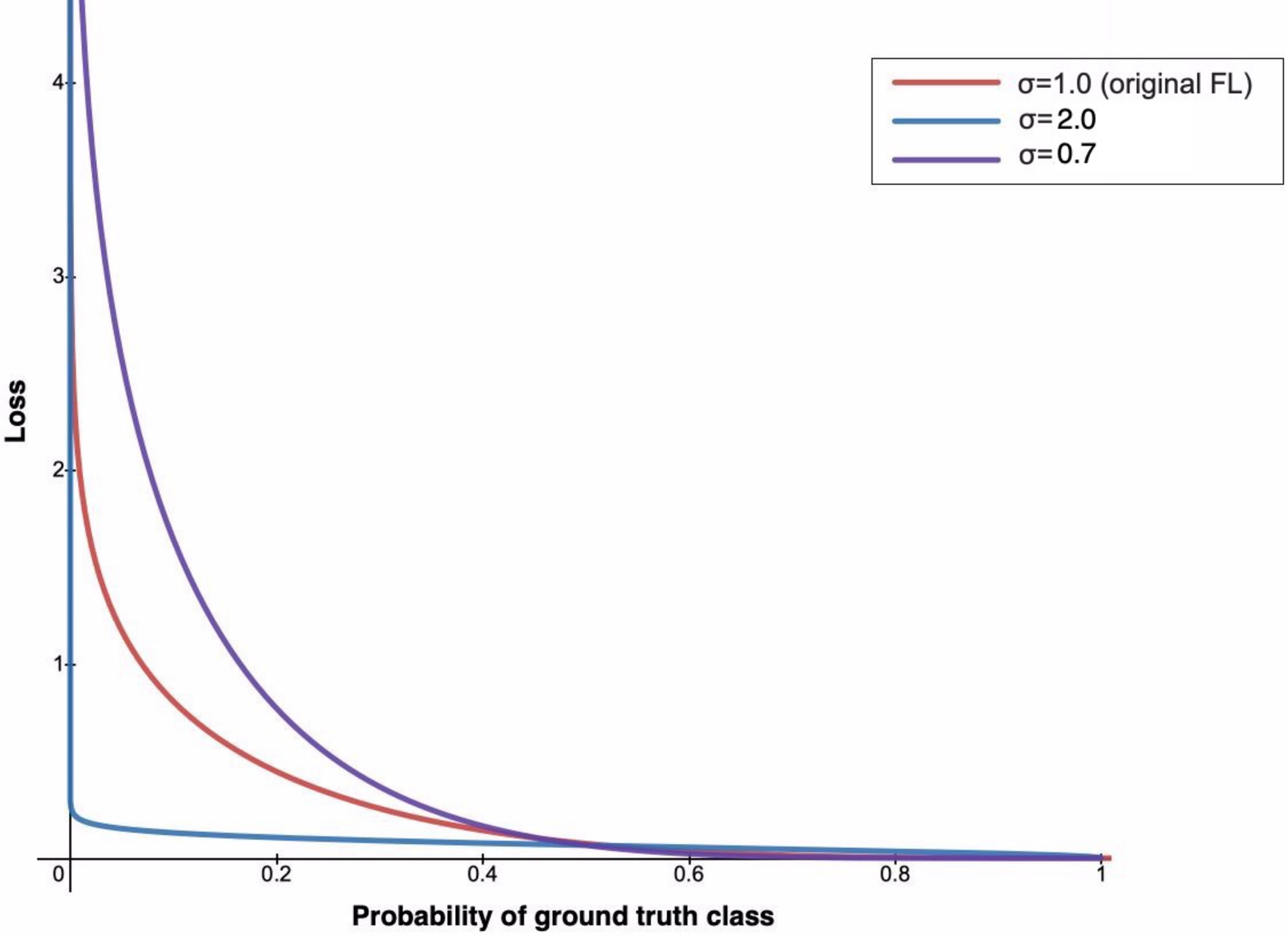}
        \caption{The proposed Bayesian Focal loss function for different estimates of aleatoric uncertainty at $\sigma=0.7$, $\sigma=1.0$, $\sigma=2.0$. At $\sigma=1.0$ function is equal to the original Focal loss.}\label{fig:bfl}
\end{figure}

In our experiments, we train the network to predict the log variance, $s:=\log\sigma^2$ to preserve the numerical stability.
The proposed Bayesian Focal loss function plot is presented in Fig.~\ref{fig:bfl} in comparison with the original Focal loss. 
Our loss function penalizes the neural network better than the original Focal loss ($\sigma$=1 in the figure): for less noisy data, it penalizes the neural network less for well-classified examples and more for large prediction errors.
For noisier data, it penalizes the neural network more uniformly, less ``trusting'' the data labels.

\subsection{Multi-task Likelihood for Bayesian RetinaNet}


For the multi-task Bayesian RetinaNet with output $y_1$ for a localization task and $y_2$ for a classification task, we obtain the following minimization objective:
\begin{equation}
\begin{aligned}
    L(W, \sigma_1,\sigma_2)&= BSmooth_{L1}\left(f^W(x),\sigma_1\right) +\\
    &+ \alpha \cdot BFL\left(f^W(x),\sigma_2\right)
\end{aligned}
\end{equation}
where $BSmooth_{L1}\left(f^W(x),\sigma_1\right)$ is the Bayesian Smooth $L_1$ loss for $y_1$, $BFL\left(f^W(x),\sigma_2\right)$ is the Bayesian Focal loss for $y_2$, $\alpha$ is the balancing coefficient that adjusts the impact of the $BFL\left(f^W(x),\sigma_2\right)$ term. 
This multi-task loss is optimised with respect to $W$ as well as $\sigma_1$ and $\sigma_2$.

Unlike in~\citep{yarin-gal}, our multi-task objective does not allow to weigh losses by tuning $\sigma_1$ and $\sigma_2$. 
It only allows to learn these noise scalars and thus capture homoscedastic uncertainty.

\section{Experiments and Results}\label{sec:exp}

In our experiments, as the backbone for RetinaNet and Bayesian RetinaNet we used only ResNet-50 due to the memory limitation. 
The architecture of Bayesian RetinaNet was the same as the original RetinaNet model. 
The changes were made only for losses, which were replaced with the developed objectives. 
For both models, we used image scale equal to 800.

\subsection{Dataset}
We evaluated our loss functions on the COCO 2017 dataset~\citep{coco}. This dataset is known for being quite noisy~\citep{khetan2017learning,vahdat2017toward}, because it was crapped from the Flickr image database.
The dataset consists of more than 330,000 images, with 220,000 labelled images and more than 1.5 million objects in total. 
All objects are presented in the wild. 
The COCO dataset contains 80 object classes. 
Commonly, images contain objects of multiple classes, but about 10\% contain a single class only. 
All objects are annotated with bounding box coordinates and classes, which are stored in the JSON format. 

\subsection{Evaluation}
Experiments were conducted on a single NVIDIA Titan RTX GPU with 24GB of VRAM. 
The original implementation of the RetinaNet model was taken from the \texttt{detectron2}~\citep{detectron2} library, based on the \texttt{pytorch}~\citep{pytorch} framework. 

First, we trained the original RetinaNet model using Adam~\citep{adam} optimizer with an initial learning rate of 0.00001. The learning rate scheduler with warmup was used.

Next, we trained our model, which is Bayesian RetinaNet, using Adam optimizer with an initial learning rate of 0.00001. The learning rate scheduler with warmup was also used. We initialized $s_1=\log\sigma_1^2$ for the localization task with 1.0, $s_2=\log\sigma_2^2$ for the classification task with 0.0. Both models training took 900,000 iterations, which is about 3 days on average. For our model we conducted 5-fold cross-validation. 

For evaluation, we used a standard script from the \texttt{cocoapi}~\citep{coco} library. 
Models were evaluated on the \texttt{val} and \texttt{test-dev} splits of the MS COCO 2017 dataset. 
The primary metric of the COCO is mean average precision (\textit{mAP}).

\subsection{Results Analysis}
Tables~\ref{tab3:results_val} and~\ref{tab3:results_test_dev} show the results of comparing metrics for our model and the original RetinaNet-ResNet-50 model. The original model achieved 35.9\% mAP on the val set and 35.7\% mAP on the test-dev set, as reported in paper~\citep{focal-loss}.
\begin{table}[!h]
    \caption{Comparison of RetinaNet trained with original loss functions and Bayesian RetinaNet trained with proposed loss functions, which model homoscedastic aleatoric uncertainty, on the \texttt{val} set of the MS COCO dataset. Here, mAP is mean average precision presented for different IoU thresholds and object sizes (small, medium, large), mAR is mean average recall presented for different numbers of detections per image and object sizes. The results of Bayesian RetinaNet are presented with a standard deviation.}
    \label{tab3:results_val}
    \centering
    \begin{tabular}{lcc}
        \toprule 
        \bfseries Metric & \bfseries RetinaNet & \bfseries Bayesian RetinaNet (our) \\\hline
        \midrule 
        mAP & 35.9\% & \textbf{37.0$\pm$0.2\%} \\
        mAP50  & 54.2\% & \textbf{55.2$\pm$0.6\%} \\
        mAP75 & 38.4\% & \textbf{39.7$\pm$0.4\%} \\
        mAPs  & 20.6\% &  \textbf{21.1$\pm$0.1\%} \\
        mAPm  & 38.9\% & \textbf{40.6$\pm$0.4\%} \\
        mAPl  & 46.2\% & \textbf{47.7$\pm$0.7\%} \\
        mARmax1 & 31.8\% & \textbf{32.2$\pm$0.3\%} \\
        mARmax10 & 51.6\% &  \textbf{51.9$\pm$0.6\%} \\
        mARmax100 & 54.8\% &  \textbf{55.1$\pm$0.7\%} \\
        mARs  & \textbf{35.8\%} & 35.2$\pm$1.1\% \\
        mARm  & 58.5\% & \textbf{59.2$\pm$0.5\%} \\
        mARl  & 69.2\% &  \textbf{69.8$\pm$0.8\%} \\
        \bottomrule 
    \end{tabular}
\end{table}

\begin{table}[!h]
    \caption{Comparison of RetinaNet trained with original loss functions and Bayesian RetinaNet trained with proposed loss functions, which model homoscedastic aleatoric uncertainty, on the \texttt{test-dev} set of the MS COCO dataset. Here, mAP is mean average precision presented for different IoU thresholds and object sizes (small, medium, large). The results of Bayesian RetinaNet are presented with a standard deviation.}
    \label{tab3:results_test_dev}
    \centering
    \begin{tabular}{lcc}
        \toprule 
        \bfseries Metric & \bfseries RetinaNet & \bfseries Bayesian RetinaNet (our) \\\hline
        \midrule 
        mAP & 35.7\% & \textbf{37.4$\pm$0.1\%} \\
        mAP50  & 55.0\% & \textbf{55.7$\pm$0.5\%} \\ 
        mAP75 & 38.5\% & \textbf{40.2$\pm$0.2\%} \\
        mAPs  & 18.9\% & \textbf{21.1$\pm$0.3\%} \\
        mAPm  & 38.9\% & \textbf{39.8$\pm$0.2\%} \\
        mAPl  & 46.3\% & \textbf{46.5$\pm$0.4\%} \\
        \bottomrule 
    \end{tabular}
\end{table}

As can be seen, our model provides an average increase of 1.7\% for the main mAP metric on the test-dev set and increase of 1.1\% on the val set. 
This result seems to confirm the hypothesis that modeling aleatoric uncertainty can improve the accuracy of the detection problem solving, which answers the \textbf{RQ1}. 
We can conclude that our proposed losses penalize the neural network better than the original losses of RetinaNet. 
The average estimations of aleatoric uncertainties obtained during the training were $0.124$ for the regression task and $0.805$ for the classification task. 
These values correlate with the fact that the COCO dataset has noisy labels. 

While all the average precision metrics obtained by Bayesian RetinaNet are higher or equal compared to the baseline, average recall metrics on the val set are better only in 2 of 5 cases. 
The reason of such an effect is that our loss functions penalize the model more for false positive errors, while the true positive rate increases less significantly. 
This fact is consistent with the functions plots: for example, Bayesian Focal loss provides higher gradient values for bigger errors than the original Focal loss. 

The proposed loss functions are easy for incorporating to the existing neural networks, that utilize cross-entropy/Focal loss for the classification and $L_1$/Smooth $L_1$ loss for the localization tasks solving. 
Thus, in future our losses can be scaled and applied for SpineNet~\citep{spinenet}, ATSS~\citep{atss} and other current state-of-the-art detection models. This answers \textbf{RQ2}. 
Modeling homoscedastic aleatoric uncertainty can advance the neural network detectors robustness, help them better generalize to the real-world scenarios and achieve higher performance.

\section{Conclusion and Discussion}\label{sec:outro}

In this work, we have proposed the novel loss functions for the detection problem (i.e. joint classification and localization), namely \textit{Bayesian Focal loss} and \textit{Bayesian Smooth} $L_1$ \textit{loss} functions. 
The proposed functions are able to model homoscedastic aleatoric uncertainty during model training and do not require the architecture changes.

The proposed losses were studied using the COCO 2017 dataset based on the RetinaNet-ResNet-50 model. As a result of the study, an increase of 1.7\% by the mAP metric on the \texttt{test-dev} set was achieved. The obtained result confirms the hypothesis that modeling homoscedastic aleatoric uncertainty improves the accuracy of the detection problem solution. 
The average values of aleatoric uncertainties obtained using our losses were $0.805$ for the classification task and $0.124$ for the regression task.

In future work, we plan to apply the proposed loss functions to other models, which are based on the RetinaNet architecture, for example, SpineNet~\citep{spinenet}, ATSS~\citep{atss}. 
We also plan to evaluate the developed functions on other datasets with known noise values to prove that the uncertainties estimates correlate with these values. Furthermore, it is interesting to apply the developed loss functions to model heteroscedastic aleatoric uncertainty, which can advance the detection accuracy and increase the interpretability of detection per object.

\begin{contributions} 

    N.~Khanzhina conceived the idea, developed the proposed functions, supervised the research and wrote the paper.
    A.~Lapenok helped developing the proposed functions, created the code, conducted experiments, made the figures and wrote the paper.
    A.~Filchenkov consulted on mathematical background and wrote the paper.
\end{contributions}

\begin{acknowledgements}  
This work is financially supported by National Center for Cognitive Research of ITMO University.

The authors would like to thank Tatyana Polevaya, and Georgy Zamorin for their great help, Evgeny Tsymbalov, Alex Farseev, and Inna Anokhina for useful comments.
\end{acknowledgements}



\end{document}